\documentclass{article}
\usepackage[utf8]{inputenc}

\usepackage{parskip}
\usepackage[natbibapa]{apacite}
\usepackage[title]{appendix}

\usepackage{siunitx}
\usepackage{eurosym}

\usepackage{amsmath}

\usepackage{geometry}
\usepackage[pdftex]{graphicx}

\usepackage{comment}

\usepackage{hyperref}

\usepackage{caption}
\usepackage{multicol}
\usepackage{booktabs} 

\usepackage{enumitem}

\title{Five Points to Check when Comparing Visual Perception in Humans and Machines}

\author{Christina M. Funke*\textsuperscript{1}, Judy Borowski*\textsuperscript{1}, Karolina Stosio\textsuperscript{1-3}, \\Wieland Brendel\textsuperscript{$\dagger$ 1, 2, 4}, Thomas S. A. Wallis\textsuperscript{$\dagger$ 1, 5}, Matthias Bethge\textsuperscript{$\dagger$ 1, 2, 4}\\
\small{* joint first authors, $\dagger$ joint senior authors} \\
\small{\textsuperscript{1} University of T\"ubingen, Germany}\\
\small{\textsuperscript{2}Bernstein Center for Computational Neuroscience, T\"ubingen and Berlin, Germany}\\
\small{\textsuperscript{3}Volkswagen Group Machine Learning Research Lab, Munich, Germany}\\
\small{\textsuperscript{4}Werner Reichardt Centre for Integrative Neuroscience, T\"ubingen, Germany}\\
\small{\textsuperscript{5}Current affiliation Amazon.com, T\"ubingen; this contribution is prior work.}\\
\small{corresponding authors: christina.funke@bethgelab.org and judy.borowski@bethgelab.org}
}
\date{}

\begin{document}
\sisetup{range-phrase=--}

\maketitle
\begin{abstract}
With the rise of machines to human-level performance in complex recognition tasks, a growing amount of work is directed towards comparing information processing in humans and machines. These studies are an exciting chance to learn about one system by studying the other. Here, we propose ideas on how to design, conduct and interpret experiments such that they adequately support the investigation of mechanisms when comparing human and machine perception.
We demonstrate and apply these ideas through three case studies.
The first case study shows how human bias can affect the interpretation of results and that several analytic tools can help to overcome this human reference point.
In the second case study, we highlight the difference between necessary and sufficient mechanisms in visual reasoning tasks. Thereby, we show that contrary to previous suggestions, feedback mechanisms might not be necessary for the tasks in question.
The third case study highlights the importance of aligning experimental conditions. We find that a previously-observed difference in object recognition does not hold when adapting the experiment to make conditions more equitable between humans and machines.
In presenting a checklist for comparative studies of visual reasoning in humans and machines, we hope to highlight how to overcome potential pitfalls in design or inference.

\end{abstract}

\begin{quote}
\end{quote}

\section*{Introduction}

Until recently, only biological systems could abstract the visual information in our world and transform it into a representation that supports understanding and action. Researchers have been studying how to implement such transformations in artificial systems since at least the 1950s. One advantage of artificial systems for understanding these computations is that many analyses can be performed that would not be possible in biological systems. For example, key components of visual processing, such as the role of feedback connections, can be investigated, and methods such as ablation studies gain new precision.

\begin{sloppypar}
Traditional models of visual processing sought to explicitly replicate the hypothesized computations performed in biological visual systems. One famous example is the hierarchical HMAX-model \citep{fukushima1980neocognitron, riesenhuber1999hierarchical}. It instantiates mechanisms hypothesized to occur in primate visual systems, such as template matching and max operations, whose goal is to achieve invariance to position, scale and translation. Crucially, though, these models never got close to human performance in real-world tasks.
\end{sloppypar}

With the success of learned approaches in the last decade, and particularly that of convolutional deep neural networks (DNNs), we now have much more powerful models. In fact, these models are able to perform a range of constrained image understanding tasks with human-like performance \citep{krizhevsky2012imagenet, eigen2015predicting, long2015fully}.

While matching machine performance with that of the human visual system is a crucial step, the inner workings of the two systems can still be very different. We hence need to move beyond comparing accuracies to understand how the systems' mechanisms differ \citep{geirhos2020beyond, chollet2019measure, ma2020neural, Firestone201905334}.

The range of frequently considered mechanisms is broad. They concern not only the architectural level (such as feedback vs. feed-forward connections, lateral connections, foveated architectures or eye movements, ...), but also involve different learning schemes (Back-propagation vs Spike-timing-dependent plasticity/Hebbian learning, ...) as well as the nature of the representations themselves (such as reliance on texture rather than shape, global vs. local processing, ...)\footnote{For an overview of comparison studies, please see Appendix \ref{related_work}}.

\section*{Checklist for Psychophysical Comparison Studies}
\begin{figure}
 \centering 
 \includegraphics[width=\linewidth]{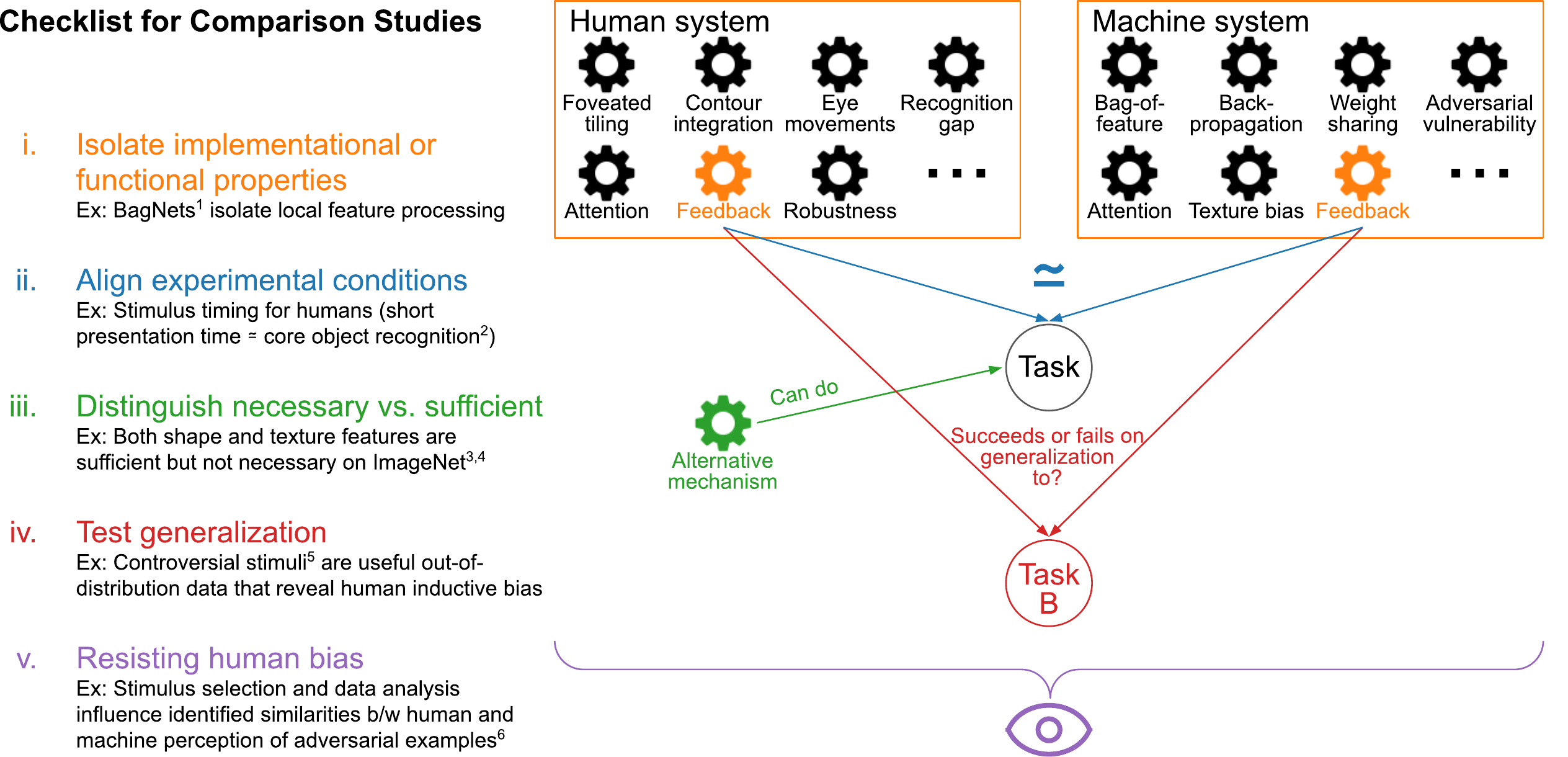} 
 \caption{\textbf{i}: The human system and a candidate machine system differ in a range of properties. Isolating a specific mechanism (for example feedback) can be challenging. \textbf{ii}: When designing an experiment, equivalent settings are important. \textbf{iii}: Even if a specific mechanism was important for a task, it would not be clear if this mechanism is necessary, as there could be other mechanisms (that might or might not be part of the human or machine system) that can allow a system to perform well. \textbf{iv}: Furthermore, the identified mechanisms might depend on the specific experimental setting and not generalize to e.g. another task. \textbf{v}: Overall, our human bias influences how we conduct and interpret our experiments. \textsuperscript{1}\citet{brendel2019approximating}, \textsuperscript{2}\citet{dicarlo2012does}, \textsuperscript{3}\citet{geirhos2018imagenet}, \textsuperscript{4}\citet{kubilius2016deep}, \textsuperscript{5}\citet{golan2019controversial}, \textsuperscript{6}\citet{dujmovic2020adversarial}
 }
 \label{fig:framework_figure}
\end{figure}

We present a checklist on how to design, conduct and interpret experiments of comparison studies that investigate relevant mechanisms for visual perception. The diagram in Figure~\ref{fig:framework_figure} illustrates the core ideas which we elaborate on below.

\begin{enumerate}[label=\roman*.]

    \item \textbf{Isolating implementational or functional properties.} Naturally, the systems that are being compared often differ in more than just one aspect, and hence pinpointing one single reason for an observed difference can be challenging. One approach is to design an artificial network constrained such that the mechanism of interest will show its effect as clearly as possible. An example of such an attempt is \citet{brendel2019approximating} which constrained models to process purely local information by reducing their receptive field sizes. Unfortunately, in many cases, it is almost impossible to exclude potential side-effects from other experimental factors such as architecture or training procedure. Therefore, making explicit if, how and where results depend on other experimental factors is important.
       
    \item \textbf{Aligning experimental conditions for both systems.} In comparative studies (whether humans and machines, or different organisms in nature), it can be exceedingly challenging to make experimental conditions equivalent. When comparing the two systems, any differences should be made as explicit as possible and taken into account in the design and analysis of the study. For example the human brain profits from lifelong experience, whereas a machine algorithm is usually limited to learning from specific stimuli of a particular task and setting. Another example is the stimulus timing used in psychophysical experiments, for which there is no direct equivalent in stateless algorithms. Comparisons of human and machine accuracies must therefore be considered with the temporal presentation characteristics of the experiment. These characteristics could be chosen based on, for example, a definition of the behaviour of interest as that occurring within a certain time after stimulus onset \citep[as for e.g. ``core object recognition'';][]{dicarlo2012does}.
    \citet{Firestone201905334} highlights that as aligning systems perfectly may not be possible due to different ``hardware'' constraints such as memory capacity, unequal performance of two systems might still arise despite similar competencies.
        
    \item \textbf{Differentiating between necessary and sufficient mechanisms.} It is possible that multiple mechanisms allow good task performance  -- for example DNNs can use either shape or texture features to reach high performance on ImageNet \citep{geirhos2018imagenet, kubilius2016deep}. Thus, observing good performance for one mechanism does neither imply that this mechanism is strictly necessary nor that it is employed by the human visual system. 
    As another example, \cite{watanabe2018illusory} investigated whether the rotating snakes illusion \citep{kitaoka2003phenomenal, conway2005neural} could be replicated in artificial neural networks. While they found that this was indeed the case, we argue that the mechanisms must be different from the ones used by humans, as the illusion requires small eye movements or blinks \citep{hisakata2008effects, kuriki2008functional}, while the artificial model does not emulate such biological processes.
    
    \item \textbf{Testing generalization of mechanisms.} Having identified an important mechanism, one needs to make explicit for which particular conditions (class of tasks, data sets, ...) the conclusion is intended to hold. A mechanism that is important for one setup may or may not be important for another one. In other words, whether a mechanism works under generalized settings has to be explicitly tested.
    An example of outstanding generalization for humans is their visual \emph{robustness} against various variations in the input. In DNNs, a mechanism to improve robustness is to ``stylize'' \citep{gatys2016image} training data. First presented as raising performance on parametrically distorted images \citep{geirhos2018imagenet}, this mechanism was later shown to also improve performance on images suffering from common corruptions \citep{michaelis2019benchmarking}, but would be unlikely to help with adversarial robustness.
    From a different perspective, the work of \citet{golan2019controversial} on controversial stimuli is an example where using stimuli outside of the training distribution can be insightful. Controversial stimuli are synthetic images that are designed to trigger distinct responses for two machine models. In their experimental setup, the use of this out-of-distribution data allows the authors to reveal whether the inductive bias of humans is similar to one of the candidate models.
    
    \item \textbf{Resisting human bias.} 
    Human bias can affect not only the design but also the conclusions we draw from comparison experiments. In other words, our human reference point can influence for example how we interpret the behaviour of other systems, be they biological or artificial.
    An example is the well-known Braitenberg vehicles \citep{braitenberg1986vehicles}, which are defined by very simple rules. To a human observer, however, the vehicles' behaviour appears as arising from complex internal states such as fear, aggression or love. This phenomenon of anthropomorphizing is well known in the field of comparative psychology \citep{romanes1883animal, wolfgang1925mentality, koehler1943zahl, haun2011origins, boesch2007makes, tomasello2008assessing}. \citet{buckner2019comparative} specifically warns of human-centered interpretations and recommends to apply the lessons learned in comparative psychology to comparing DNNs and humans.
    In addition, our human reference point can influence how we design an experiment. As an example, \citet{dujmovic2020adversarial} illustrate that the selection of stimuli and labels can have a big effect on finding similarities or differences between humans and machines to adversarial examples.

\end{enumerate}

In the remainder of this paper, we provide concrete examples of the aspects discussed above using three case studies\footnote{The code is available at \url{https://github.com/bethgelab/notorious_difficulty_of_comparing_human_and_machine_perception}}:

\begin{enumerate}
    \item \textbf{Closed Contour Detection}: The first case study illustrates how tricky overcoming our human bias can be, and that shedding light on an alternative decision-making mechanism may require multiple additional experiments. 
    \item \textbf{Synthetic Visual Reasoning Test}: The second case study highlights the challenge of isolating mechanisms and of differentiating between necessary and sufficient mechanisms. Thereby, we discuss how human and machine model learning differ and how changes in the model architecture can affect the performance.
    \item \textbf{Recognition gap}: The third case study illustrates the importance of aligning experimental conditions.
\end{enumerate}

\section*{Case Study 1: Closed Contour Detection}

Closed contours play a special role in human visual perception. According to the Gestalt principles of pr\"agnanz and good continuation, humans can group distinct visual elements together so that they appear as a ``form'' or ``whole''. As such, closed contours are thought to be prioritized by the human visual system and to be important in perceptual organization \citep{koffka2013principles, elder1993effect, kovacs1993closed, tversky2004contour, ringach1996spatial}. Specifically, to tell if a line closes up to form a closed contour, humans are believed to implement a process called ``contour integration'' that relies at least partially on global information \citep{levi2007global, loffler2003local, mathes2007closure}. Even many flanking, open contours would hardly influence human's robust closed contour detection abilities. 

\subsection*{Our Experiments}
We hypothesize that, in contrast to humans, closed contour detection is difficult for DNNs. The reason is that this task would presumably require long-range contour integration, but DNNs are believed to process mainly local information \citep{geirhos2018imagenet, brendel2019approximating}.
Here, we test how well humans and neural networks can separate closed from open contours. To this end, we create a custom data set, test humans and DNNs on it and investigate the decision-making process of the DNNs.

\subsubsection*{DNNs and Humans Reach High Performance}
\begin{figure}
 \centering 
 \includegraphics[width=\linewidth]{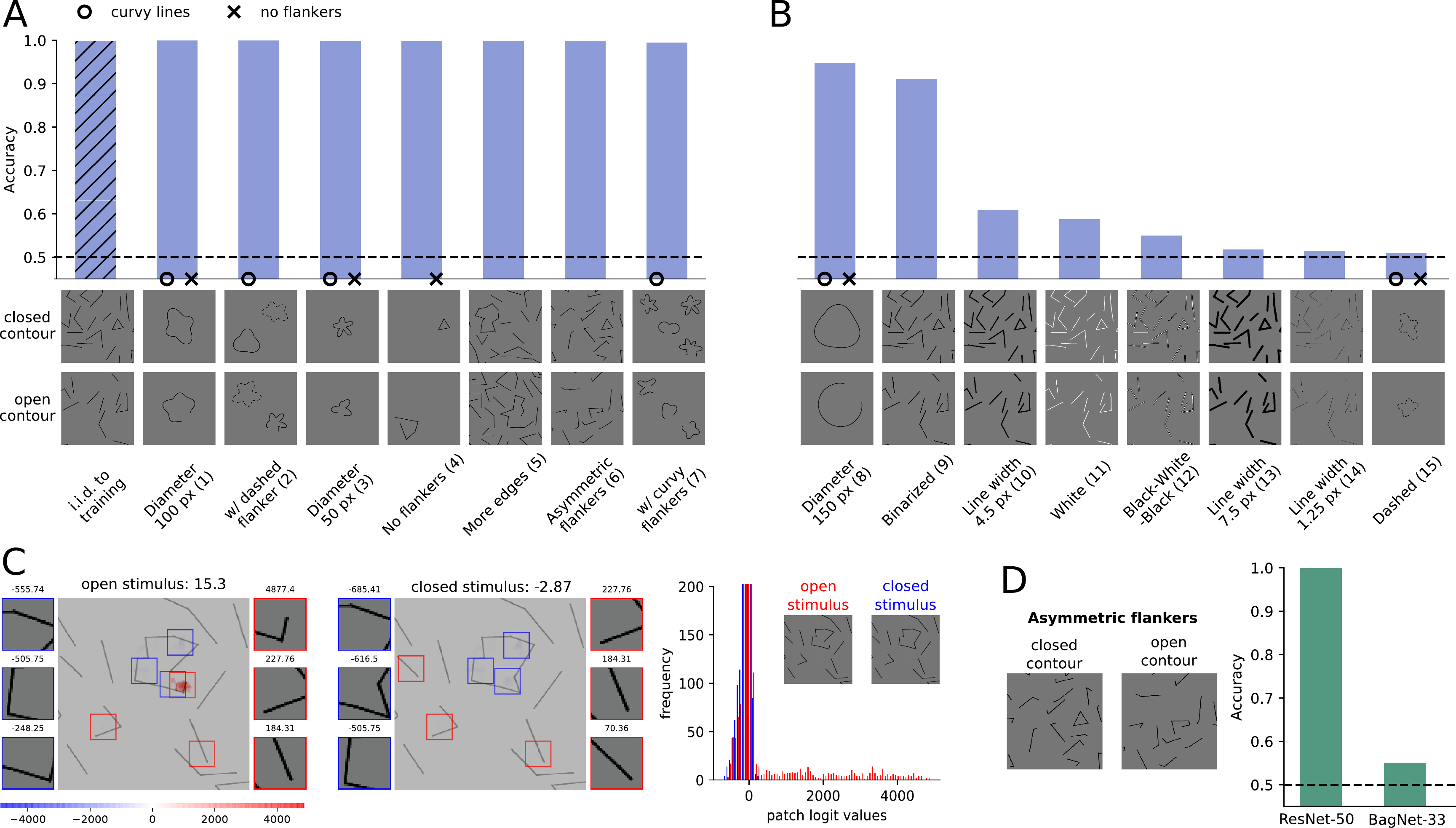} 
 \caption{
 \textbf{A}: Our ResNet-50-model generalized well to many data sets without further retraining, suggesting it would be able to distinguish closed and open contours. \textbf{B}: However, the poor performance on many other data sets showed that our model did \textit{not} learn the concept of closedness. \textbf{C}: The heatmaps of our BagNet-33-based model show which parts of the image provided evidence for closedness (blue, negative values) or openness (red, positive values). The patches on the sides show the most extremely, non-overlapping patches and their logit-values. The logit distribution shows that most patches had logit values close to zero (y-axis truncated) and that many more patches in the open stimulus contributed positive logit values. \textbf{D}: Our BagNet- and ResNet-models showed different performances on generalization sets, such as the asymmetric flankers. This indicates that the local decision-making process of the substitute model BagNet is not used by the original model ResNet. Figure best viewed electronically.}
 \label{fig:cc_main}
\end{figure}

We created a data set with two classes of images: The first class contained a closed contour, the second one did not. In order to make sure that the statistical properties of the two classes were similar, we included a main contour for both classes. While this contour line closed up for the first class, it remained open for the second class. This main contour consisted of $3 - 9$ straight line segments. In order to make the task more difficult, we added several flankers with either one or two line segments that each had a length of at least $32$ px (Figure \ref{fig:cc_main}A). The size of the images was $256 \times 256$ px. All lines were black and the background was uniformly gray. Details on the stimulus generation can be found in Appendix \ref{cc_dataset_details}.

Humans identified the closed contour stimulus very reliably in a two-interval forced choice task.
Their performance was $88.39\%$ (SEM = $2.96\%$) on stimuli whose generation procedure was identical to the training set. 
For stimuli with white instead of black lines, human participants reached a performance of $90.52\%$ (SEM = $1.58\%$). 
The psychophysical experiment is described in Appendix \ref{psychopyhysics_cc}.

We fine-tuned a ResNet-50 \citep{he2016deep} pre-trained on ImageNet \citep{imagenet_cvpr09} on the closed contour data set. Similar to humans, it performed very well and reached an accuracy of $99.95\%$ (see Figure \ref{fig:cc_main}A [i.i.d. to training]).

We found that both humans and our DNN reach high accuracy on the closed contour detection task. From a human-centered perspective it is enticing to infer that the model had learned the concept of open and closed contours and possibly that it performs a similar contour integration-like process as humans. However, this would have been overhasty. To better understand the degree of similarity, we investigated how our model performs on variations of the data sets that were not used during the training procedure.

\subsubsection*{Generalization Tests Reveal Differences}
Humans are expected to have no difficulties if the number of flankers, the color or the shape of lines would differ. We here test our model's robustness on such variants of the data set. If our model used similar decision-making processes as humans, it should be able to generalize well without any further training on the new images. This procedure is another perspective to shed light on whether our model really understood the concept of closedness or just picked up some statistical cues in the training data set.

We tested our model on $15$ variants of the data set (out of distribution test sets) without fine-tuning on these variations. As shown in Figure \ref{fig:cc_main}A and B, our trained model generalized well to many but not all modified stimulus sets. 

On the following variations, our model achieved high accuracy: 
Curvy contours ($1$, $3$) were easily distinguishable for our model, as long as the diameter remained below \SI{100}{px}.
Also, adding a dashed, closed flanker ($2$) did not lower its performance.
The classification ability of the model remained similarly high for the no flankers ($4$) and the asymmetric flankers condition ($6$).
When testing our model on main contours that consisted of more edges than the ones presented during training ($5$), the performance was also hardly impaired.
It remained high as well when multiple curvy open contours were added as flankers ($7$).

The following variations were more difficult for our model:
If the size of the contour got too large, a moderate drop in accuracy was found ($8$).
For binarized images, our model's performance was also reduced ($9$).
And finally, (almost) chance performance was observed when varying the line width ($14$, $10$, $13$), changing the line color ($11$, $12$) or using dashed curvy lines ($15$).

While humans would perform well on all variants of the closed contour data set, the failure of our model on some generalization tests suggests that it solves the task differently from humans. On the other hand it is equally difficult to prove that the model does not understand the concept. As described by \citet{Firestone201905334} models can "perform differently despite similar underlying competences". In either way, we argue that it is important to openly consider alternative mechanisms to the human approach of global contour integration.

\subsubsection*{Our Closed Contour Detection Task is Partly Solvable with Local Features}
In order to investigate an alternative mechanism to global contour integration, we here design an experiment to understand how well a decision-making process based on purely local features can work. For this purpose, we trained and tested BagNet-33 \citep{brendel2019approximating}, a model that has access to local features only. It is a variation of ResNet-50 \citep{he2016deep} where most $3 \times 3$ kernels are replaced by $1 \times 1$ kernels and therefore the receptive field size at the top-most convolutional layer is restricted to $33 \times 33$ pixels.

We found that our restricted model still reached close to $90\%$ performance. In other words, contour integration was not necessary to perform well on the task.

To understand which local features the model relied on mostly, we analyzed the contribution of each patch to the final classification decision. To this end, we used the log-likelihood values for each $33 \times 33$ pixels patch from BagNet-33 and visualized them as a \textit{heatmap}. Such a straight-forward interpretation of the contributions of single image patches is not possible with standard DNNs like ResNet \citep{he2016deep} due to their large receptive field sizes in top layers.

The heatmaps of BagNet-33 (see Figure \ref{fig:cc_main}C) revealed which local patches played an important role in the decision-making process: An open contour was often detected by the presence of an end-point at a short edge. Since all flankers in the training set had edges larger than $33$ pixels, the presence of this feature was an indicator of an open contour. In turn, the absence of this feature was an indicator of a closed contour.

Whether the ResNet-50-based model used the same local feature as the substitute model was unclear. To answer this question, we tested BagNet on the previously mentioned generalization tests. We found that the data sets on which it showed high performance were sometimes different from the ones of ResNet (see Figure \ref{fig:cc_bagnet_gen} in the Appendix). A striking example was the failure of BagNet on the "asymmetric flankers" condition (see Figure \ref{fig:cc_main}D). For these images, the flankers often consisted of shorter line segments and thus obscured the local feature we assumed BagNet to use. In contrast, ResNet performed well on this variation. This suggests that the decision-making strategy of ResNet did not heavily depend on the local feature found with the substitute BagNet model. 

In summary, the generalization tests, the high performance of BagNet as well as the existence of a distinctive local feature provide evidence that our human-biased assumption was misleading. We saw that other mechanisms for closed contour detection besides global contour integration do exist (see Introduction \textit{"Differentiating between necessary and sufficient mechanisms"}). As humans, we can easily miss the many statistical subtleties by which a task can be solved. In this respect, BagNets proved to be a useful tool to test a purportedly ``global'' visual task for the presence of local artifacts. Overall, various experiments and analyses can be beneficial to understand mechanisms and to overcome our human reference point.

\section*{Case Study 2: Synthetic Visual Reasoning Test}

In order to compare human and machine performance at learning abstract relationships between shapes, \citet{fleuret2011comparing} created the Synthetic Visual Reasoning Test (SVRT) consisting of 23 problems (see Figure \ref{fig:svrt_main}A). They showed that humans need only few examples to understand the underlying concepts.
\citet{stabinger201625} as well as \citet{kim2018not} assessed the performance of deep convolutional neural networks on these problems.
Both studies found a dichotomy between two task categories: While high accuracy was reached on spatial problems, the performance on same-different problems was poor. 
In order to compare the two types of tasks more systematically, \citet{kim2018not} developed a parameterized version of the SVRT data set called PSVRT. Using this data set, they found that for same-different problems, an increase in the complexity of the data set could quickly strain their models. In addition, they showed that an attentive version of the model did not exhibit the same deficits. From these results the authors concluded that feedback mechanisms as present in the human visual system such as attention, working memory or perceptual grouping are probably important components for abstract visual reasoning. 
More generally, these papers have been perceived and cited with the broader claim of feed-forward DNNs not being able to learn same-different relationships between visual objects \citep{serre2019deep, schofield2018understanding} - at least not ``efficiently'' \citet{Firestone201905334}.

We argue that the results of \citet{kim2018not} cannot be taken as evidence for the importance of feedback components for abstract visual reasoning:
\begin{enumerate}
    \item While their experiments showed that same-different tasks are harder to \emph{learn} for their models, this might also be true for the human visual system. Normally-sighted humans have experienced lifelong visual input; only looking at human performance with this extensive learning experience cannot reveal differences in learning difficulty. 
    \item Even if there is a difference in learning complexity, this difference is not necessarily due to differences in the inference mechanism (e.g. feed-forward vs feedback)---the large variety of other differences between biological and artificial vision systems could be critical causal factors as well.
    \item In the same line, small modifications in the learning algorithm or architecture can significantly change learning complexity. For example, changing the network depth or width can greatly improve learning performance \citep{tan2019efficientnet}.
    \item Just because a attentive version of the model can learn both types of tasks does not prove that feedback mechanisms are necessary for these tasks (see introduction: \textit{"Differentiating between necessary and sufficient mechanisms"}).
\end{enumerate}

Determining the necessity of feedback mechanisms is especially difficult because feedback mechanisms are not clearly distinct from purely feed-forward mechanisms. In fact, any finite-time recurrent network can be unrolled into a feed-forward network \citep{liao2016bridging, van2020going}.

For these reasons, we argue that the importance of feedback mechanisms for abstract visual reasoning remains unclear. 

In the following paragraph we present our own experiments on the SVRT data set and show that standard feed-forward DNNs can indeed perform well on same-different tasks. This confirms that feedback mechanisms are not strictly necessary for same-different tasks, although they helped in the specific experimental setting of \citet{kim2018not}. Furthermore, this experiment highlights that changes of the network architecture and training procedure can have large effects on the performance of artificial systems.

\subsection*{Our Experiments}

The findings of \citet{kim2018not} were based on rather small neural networks, which consisted of up to six layers. However, typical network architectures used for object recognition consist of more layers and have larger receptive fields. For this reason we tested a representative of such networks, namely ResNet-50. The experimental setup can be found in Appendix \ref{appendix_svrt}.

\begin{figure}
 \centering 
 \includegraphics[width=\linewidth]{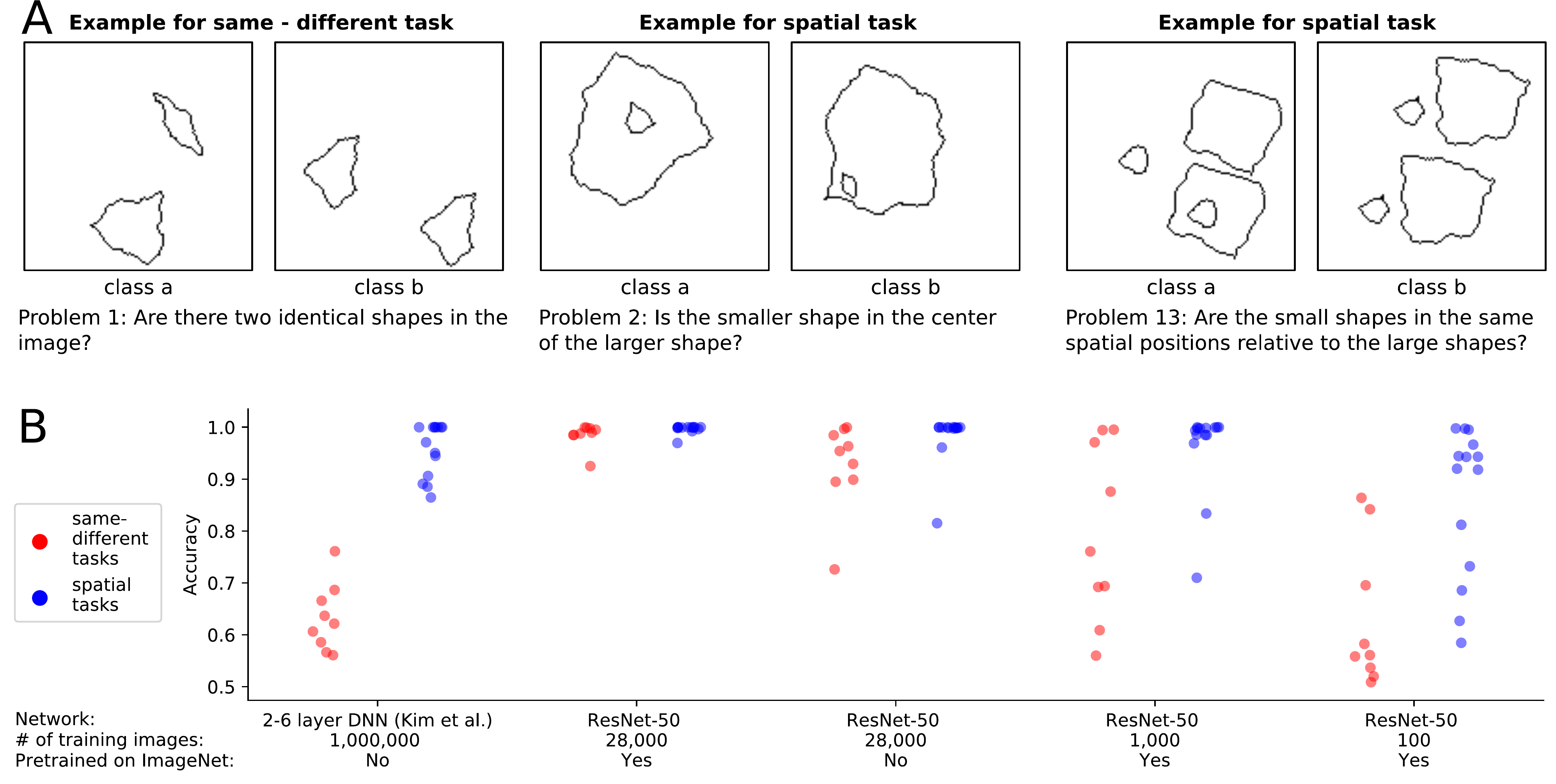}
 \caption{A: For three of the 23 SVRT problems, two example images representing the two opposing classes are shown. In each problem, the task was to find the rule that separated the images and to sort them accordingly.
 B: \citet{kim2018not} trained a DNN on each of the problems. They found that same-different tasks (red points), in contrast to spatial tasks (blue points), could not be solved with their models. Our ResNet-50-based models reached high accuracies for all problems when using $28,000$ training examples and weights from pre-training on ImageNet.}
 \label{fig:svrt_main}
\end{figure}

We found that our feed-forward model can in fact perform well on the same-different tasks of SVRT (see Figure \ref{fig:svrt_main}B, see also concurrent work of \citet{messina2019testing}). This result was not due to an increase in the number of training samples. In fact, we used fewer images ($28,000$ images) than \citet{kim2018not} ($1$ million images) and \citet{messina2019testing} (400,000 images). 
Of course, the results were obtained on the SVRT data set and might not hold for other visual reasoning data sets (see introduction \textit{"Testing generalization of mechanisms"}).

In the very low-data regime (1000 samples), we found a difference between the two types of tasks. In particular, the overall performance on same-different tasks was lower than on spatial reasoning tasks. As for the previously mentioned studies, this cannot be taken as evidence for systematic differences between feed-forward neural networks and the human visual system. In contrast to the neural networks used in this experiment, the human visual system is naturally pre-trained on large amounts of visual reasoning tasks, thus making the low-data regime an unfair testing scenario from which it is almost impossible to draw solid conclusions about differences in the internal information processing. In other words, it might very well be that the human visual system trained from scratch on the two types of tasks would exhibit a similar difference in sample efficiency as a ResNet-50. Furthermore, the performance of a network in the low-data regime is heavily influenced by many factors other than architecture, including regularization schemes or the optimizer, making it even more difficult to reach conclusions about systematic differences in the network structure between humans and machines.

\section*{Case Study 3: Recognition Gap}
\citet{ullman2016atoms} investigated the minimally necessary visual information required for object recognition. To this end, they successively cropped or reduced the resolution of a natural image until more than $50\%$ of all human participants failed to identify the object. The study revealed that recognition performance drops sharply if the minimal recognizable image crops are reduced any further. They referred to this drop in performance as the ``recognition gap''. The gap is computed by subtracting the proportion of people who correctly classify the largest unrecognizable crop (e.g. $0.2$) from that of the people who correctly classify the smallest recognizable crop (e.g. $0.9$). In this example, the recognition gap would evaluate to $0.9 - 0.2 = 0.7$. On the same human-selected crops, \citet{ullman2016atoms} found that the recognition gap is much smaller for machine vision algorithms ($0.14\pm0.24$) than for humans ($0.71\pm0.05$). The researchers concluded that machine vision algorithms would not be able to ``explain [humans'] sensitivity to precise feature configurations'' and ``that the human visual system uses features and processes that are not used by current models and that are critical for recognition''.
In a follow-up study, \citet{srivastava2019minimal} identified ``fragile recognition images'' (FRIs) with an exhaustive machine-based procedure whose results include a subset of patches that adhere to the definition of of minimal recognizable configurations (MIRCs) by \citet{ullman2016atoms}. On these machine-selected FRIs, a DNN experienced a moderately high recognition gap, whereas humans experienced a low one. Because of the differences between the selection procedures used in \citet{ullman2016atoms} and \citet{srivastava2019minimal}, the question remained open whether machines would show a high recognition gap on machine-selected minimal images, if the selection procedure was similar to the one used in \citet{ullman2016atoms}.

\subsection*{Our Experiment}
Our goal was to investigate if the differences in recognition gaps identified by \citet{ullman2016atoms} would at least in part be explainable by differences in the experimental procedures for humans and machines. Crucially, we wanted to assess machine performance on \textit{machine}-selected, and not \textit{human}-selected image crops. We therefore implemented the psychophysics experiment in a machine setting to search the smallest recognizable images (or minimal recognizable crop: `MIRCs'') and the largest unrecognizable images (``sub-MIRCs''). In the final step, we evaluated our machine model's recognition gap using the \textit{machine}-selected MIRCs and sub-MIRCs.

\paragraph{Methods}
Our machine-based search algorithm used the deep convolutional neural network BagNet-33 \citep{brendel2019approximating}, which allows to straightforwardly analyze images as small as $33 \times 33$ pixels. In the first step, the classification accuracy was evaluated for the whole image. If it was above 0.5, the image was successively cropped and reduced in resolution. In each step, the best performing crop was taken as the new parent. When the classification probability of all children fell below $0.5$, the parent was identified as the MIRC and all its children were considered sub-MIRCs.
In order to evaluate the recognition gap, we calculate the difference in accuracy between the MIRC and the \textit{best-performing} sub-MIRC. This definition is more conservative than the one from \citet{ullman2016atoms} who evaluated the difference in accuracy between the MIRC and the \textit{worst-performing} sub-MIRC. For more details on the search procedure, please see Appendix \ref{rec_gap:methods} and \ref{rec_gap:class_stride_analysis}.

\begin{figure}
 \centering 
 \includegraphics[width=\linewidth]{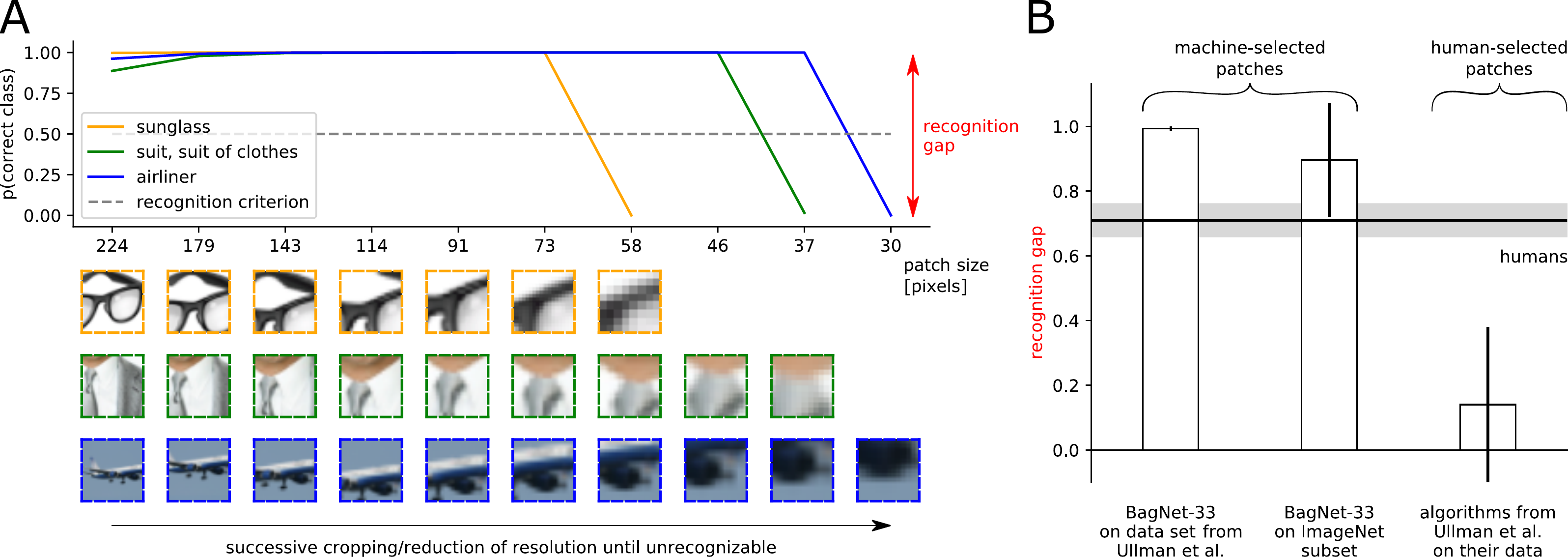}
 \caption{A: BagNet-33's probability of correct class for decreasing crops: The sharp drop when the image becomes too small or the resolution too low is called the ``recognition gap'' \citep{ullman2016atoms}. It was computed by subtracting the model's predicted probability of the correct class for the sub-MIRC from the model's predicted probability of the correct class for the MIRC. As an example, for the glasses stimulus it evaluated as $0.9999-0.0002 = 0.9997$. The crop size on the x-axis corresponds to the size of the original image in pixels. Steps of reduced resolution are not displayed such that the three sample stimuli can be displayed coherently. B: Recognition gaps for machine algorithms (vertical bars) and humans (gray horizontal bar). A recognition gap \textit{is} identifiable for the DNN BagNet-33 when testing machine-selected stimuli of the original images from \citet{ullman2016atoms} and a subset of the ImageNet validation images \citep{imagenet_cvpr09}. Error bars denote standard deviation.}
 \label{fig:recgap_main}
\end{figure}

\paragraph{Results}
We evaluated the recognition gap on two data sets: the original images from \citet{ullman2016atoms} and a subset of the ImageNet validation images \citep{imagenet_cvpr09}. As shown in Figure \ref{fig:recgap_main}A, our model has an average recognition gap of $0.99\pm0.01$ on the machine-selected crops of the data set from \citet{ullman2016atoms}. On the machine-selected crops of the ImageNet validation subset, a large recognition gap occurs as well.
Our values are similar to the recognition gap in humans and differ from the machines' recognition gap ($0.14\pm0.24$) between human-selected MIRCs and sub-MIRCs as identified by \citet{ullman2016atoms}.

\paragraph{Discussion}
Our findings contrast claims made by \citet{ullman2016atoms}. The latter study concluded that machine algorithms are not as sensitive as humans to precise feature configurations and that they are missing features and processes that are ``critical for recognition.''
First, our study shows that a machine algorithm \textit{is} sensitive to small image crops. It is only the precise minimal features that differ between humans and machines.
Second, by the word ``critical,'' \citet{ullman2016atoms} imply that object recognition would not be possible without these human features and processes. 
Applying the same reasoning to \citet{srivastava2019minimal}, the low human performance on machine-selected patches should suggest that humans would miss ``features and processes critical for recognition.'' This would be an obviously overreaching conclusion. Furthermore, the success of modern artificial object recognition speaks against the conclusion that the purported processes are ``critical'' for recognition, at least within this discretely-defined recognition task.
Finally, what we can conclude from the experiments of \citet{ullman2016atoms} and from our own is that both the human and a machine visual system \textit{can} recognize small image crops and that there \textit{is} a sudden drop in recognizability when reducing the amount of information. 

In summary, these results highlight the importance of testing humans and machines in as similar settings as possible, and of avoiding a human bias in the experiment design.
All conditions, instructions and procedures should be as close as possible between humans and machines in order to ensure that observed differences are due to inherently different decision strategies rather than differences in the testing procedure.

\section*{Conclusion}
Comparing human and machine visual perception can be challenging. In this work, we presented a checklist on how to perform such comparison studies in a meaningful and robust way.
For one, isolating a single mechanism requires us to minimize or exclude the effect of other differences between biological and artificial and to align experimental conditions for both systems. We further have to differentiate between necessary and sufficient mechanisms and to circumscribe in which tasks they are actually deployed. Finally, an overarching challenge in comparison studies between humans and machines is our strong internal human interpretation bias.

Using three case studies we illustrated the application of the checklist.
The first case study on closed contour detection showed that human bias can impede the objective interpretation of results, and that investigating which mechanisms could or could not be at work may require several analytic tools.
The second case study highlighted the difficulty of drawing robust conclusions about mechanisms from experiments. While previous studies suggested that feedback mechanisms might be important for visual reasoning tasks, our experiments showed that they are not necessarily required.
The third case study clarified that aligning experimental conditions for both systems is essential. When adapting the experimental settings, we found that, unlike the differences reported in a previous study, DNNs and humans indeed show similar behavior on an object recognition task.

Our checklist complements other recent proposals about how to compare visual inference strategies between humans and machines \citep{buckner2019comparative, chollet2019measure, ma2020neural, geirhos2020beyond} and helps to create more nuanced and robust insights into both systems.

\section*{Author contributions}
The closed contour case study was designed by CMF, JB, TSAW and MB and later with WB. The code for the stimuli generation was developed by CMF. The neural networks were trained by CMF and JB. The psychophysical experiments were performed and analysed by CMF, TSAW and JB.
The SVRT case study was conducted by CMF under supervision of TSAW, WB and MB.
KS designed and implemented the recognition gap case study under the supervision of WB and MB, JB extended and refined it under the supervision of WB and MB.
The initial idea to unite the three projects was conceived by WB, MB, TSAW and CMF, and further developed including JB. The first draft was jointly written by JB and CMF with input from TSAW and WB. All authors contributed to the final version and provided critical revisions.

\section*{Acknowledgments}
We thank Alexander S. Ecker, Felix A. Wichmann, Matthias K\"ummerer, Dylan Paiton as well as Drew Linsley for helpful discussions. We thank Thomas Serre, Junkyung Kim, Matthew Ricci, Justus Piater, Sebastian Stabinger, Antonio Rodr\'iguez-S\'anchez, Shimon Ullman, Liav Assif and Daniel Harari for discussions and feedback on an earlier version of this manuscript. Additionally, we would like to thank Nikolas Kriegeskorte for his detailed and constructive feedback, which helped us make our manuscript stronger. Furthermore, we thank Wiebke Ringels for helping with data collection for the psychophysical experiment.

We thank the International Max Planck Research School for Intelligent Systems (IMPRS-IS) for supporting CMF and JB.
We acknowledge support from the German Federal Ministry of Education and Research (BMBF) through the competence center for machine learning (FKZ 01IS18039A) and the Bernstein Computational Neuroscience Program T\"ubingen (FKZ: 01GQ1002), the German Excellence Initiative through the Centre for Integrative Neuroscience T\"ubingen (EXC307), and the Deutsche Forschungsgemeinschaft (DFG; Projektnummer 276693517 – SFB 1233).

Elements of this work were presented at the Conference on Cognitive Computational Neuroscience 2019 and the Shared Visual Representations in Human and Machine Intelligence Workshop at the Conference on Neural Information Processing Systems 2019.

\section*{Commercial relationships}
Matthias Bethge: Amazon scholar Jan 2019 – Jan 2021, Layer7AI, DeepArt.io, Upload AI; Wieland Brendel: Layer7AI.
\newpage
\bibliography{references.bib}

\newpage
\begin{appendices}
\section{Literature Overview of Comparison Studies}
\label{related_work}
A growing body of work discusses comparisons of humans and machines on a higher level. \citet{majaj2018deep} provide a broad overview how machine learning can help vision scientists to study biological vision, while \cite{barrett2019analyzing} review methods how to analyze representations of biological and artificial networks.
From the perspective of cognitive science, \cite{cichy2019deep} stress that Deep Learning models \textit{can} serve as scientific models that not only provide both helpful predictions and explanations but that can also be used for exploration.
Furthermore, from the perspective of psychology and philosophy, \cite{buckner2019comparative} emphasizes often-neglected caveats when comparing humans and DNNs such as human-centered interpretations and calls for discussions regarding how to properly align machine and human performance. \cite{chollet2019measure} proposes a general Artificial Intelligence benchmark and suggests to rather evaluate intelligence as ``skill-acquisition efficiency'' than to focus on skills at specific tasks. 

In the following, we give a brief overview of studies that compare human and machine perception. 
In order to test if DNNs have similar cognitive abilities as humans, a number of studies test DNNs on abstract (visual) reasoning tasks \citep{barrett2018measuring, yan2017intelligent, wu2019challenge, santoro2017simple, villalobos2019deep}.
Other comparison studies focus on whether human visual phenomena such as illusions \citep{gomez2019convolutional, watanabe2018illusory, kim2019neural} or crowding \citep{volokitin2017deep, doerig2019crowding} can be reproduced in computational models.
In the attempt to probe intuition in machine models, DNNs are compared to intuitive physics engines, i.e. probabilistic models that simulate physical events \citep{zhang2016comparative}.

Other works investigate whether DNNs are sensible models of human perceptual processing. To this end, their prediction or internal representations are compared to those of biological systems; for example to human and/or monkey behavioral representations \citep{peterson2016adapting, schrimpf2018brain, yamins2014performance, eberhardt2016deep, golan2019controversial}, human fMRI representations \citep{han2019representation, khaligh2014deep} or monkey cell recordings \citep{schrimpf2018brain, khaligh2014deep, yamins2014performance, cadena2019deep}.

A great number of studies focus on manipulating tasks and/or models. Researchers often use generalization tests on data dissimilar to the training set \citep{zhang2018can, wu2019challenge} to test whether machines understood the underlying concepts.
In other studies, the degradation of object classification accuracy is measured with respect to image degradations \citep{geirhos2018generalisation} or with respect to the type of features that play an important role for human or machine decision-making \citep{geirhos2018imagenet, brendel2019approximating, kubilius2016deep, ullman2016atoms, ritter2017cognitive}.
A lot of effort is being put into investigating whether humans are vulnerable to small, adversarial perturbations in images \citep{elsayed2018adversarial, zhou2019humans, han2019representation, dujmovic2020adversarial} - as DNNs are shown to be \citep{szegedy2013intriguing}. 
Similarly, in the field of Natural Language Processing, a trend is to manipulate the data set itself by for example negating statements to test whether a trained model gains an understanding of natural language or whether it only picks up on statistical regularities \citep{niven2019probing, mccoy2019right}.

Further work takes inspiration from biology or uses human knowledge explicitly in order to improve DNNs.
\citet{spoerer2017recurrent} found that recurrent connections, which are abundant in biological systems, allow for higher object recognition performance, especially in challenging situations such as in the presence of occlusions - in contrast to pure feed-forward networks.
Furthermore, several researchers suggest \citep{zhang2018can, kim2018not} or show \citep{wu2019challenge, barrett2018measuring, santoro2017simple} that designing networks' architecture or features with human knowledge is key for machine algorithms to successfully solve abstract (reasoning) tasks.

\section{Closed Contour Detection}

\subsection{Data Set} \label{cc_dataset_details}
\label{resnet_cc}
Each image in the training set contained a main contour, multiple flankers and a background image.
The main contour and flankers were drawn into an image of size $1028 \times 1028$ px. 
The main contour and flankers could either be straight or curvy lines, for which the generation processes are respectively described in \ref{polygon_desc} and \ref{curvy_desc}.
The lines had a default thickness of \SI{10}{px}.
We then re-sized the image to $256 \times 256$ px using anti-aliasing to transform the black and white pixels into smoother lines that had gray pixels at the borders. Thus, the lines in the re-sized image had a thickness of \SI{2.5}{px}.
In the following, all specifications of sizes refer to the re-sized image (i.e a line described of final length \SI{10}{px} extended over \SI{40}{px} when drawn into the $1028 \times 1028$ px image).
For the psychophysical experiments (see \ref{psychopyhysics_cc}), we added a white margin of \SI{16}{px} on each side of the image to avoid illusory contours at the borders of the image.

\paragraph{Varying Contrast of Background}
An image from the ImageNet data set was added as background to the line drawing.
We converted the image into LAB color space and linearly rescaled the pixel intensities of the image to produce a normalized contrast value between $0$ (gray image with the RGB values $[118, 118, 118]$) and $1$ (original image) (see Figure \ref{fig:cc_randomcontrast}A). 
When adding the image to the line drawing, we replaced all pixels of the line drawing by the values of the background image for which the background image had a higher grayscale value than the line drawing. 
For the experiments in the main body, the contrast of the background image was always $0$.
Only for the additional experiment described in \ref{random_contrast}, we used other contrast levels.

\paragraph{Generation of Image Pairs}\label{appendix_pairs}
We aimed to reduce the statistical properties that could be exploited to solve the task without judging the closedness of the contour. 
Therefore, we generated image pairs consisting of an "open" and a "closed" version of the same image. The two versions were designed to be almost identical and had the same flankers. They differed only in the main contour, which was either open or close.
Examples of such image pairs are shown in Figure \ref{fig:cc_methods}. 
During training, either the closed or the open image of a pair was used.
However, for the validation and testing, both versions were used. This allowed us to compare the predictions and heatmaps for images that differed only slightly, but belonged to different classes.

\subsubsection{Line-drawing with Polygons as Main Contour} \label{polygon_desc}
\begin{figure}
 \centering 
 \includegraphics[width=0.5\linewidth]{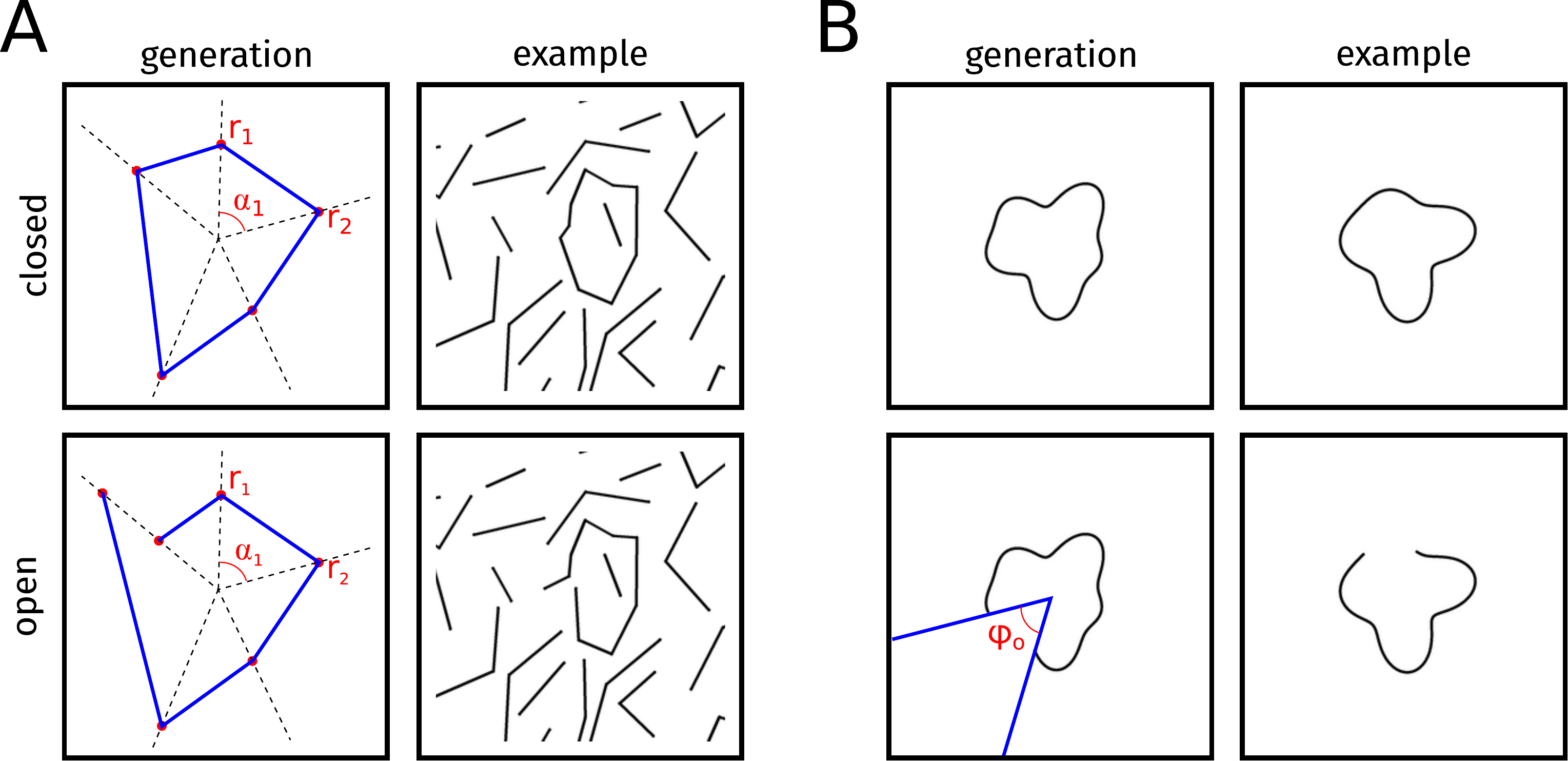}
 \caption{Closed contour data set. \textbf{A}: Left: The main contour was generated by connecting points from a random sampling process of angles and radii. Right: Resulting line-drawing with flankers. \textbf{B}: Left: Generation process of curvy contours. Right: Resulting line-drawing.}
 \label{fig:cc_methods}
\end{figure}

The data set used for training as well as some of the generalization sets consisted of straight lines.
The main contour consisted of n $\in$ \{3, 4, 5, 6, 7, 8, 9\} line segments that formed either an open or a closed contour.
The generation process of the main contour is depicted on the left side of Figure \ref{fig:cc_methods}A.
To get a contour with $n$ edges, we generated $n$ points which were defined by a randomly sampled angle $\alpha_n$ and a randomly sampled radius $r_n$ (between $0$ and \SI{128}{px}). 
By connecting the resulting points, we obtained the closed contour. 
We used the python PIL library (PIL 5.4.1, python3) to draw the lines that connect the endpoints.
For the corresponding open contour, we sampled two radii for one of the angles such that they had a distance of \SIrange{20}{50}{px} from each other.
When connecting the points, a gap was created between the points that share the same angle.
This generation procedure could allow for very short lines with edges being very close to each other.
To avoid this we excluded all shapes with corner points closer to \SI{10}{px} from non-adjacent lines.

The position of the main contour was random, but we ensured that the contour did not extend over the border of the image.

Besides the main contour, several flankers consisting of either one or two line segments were added to each stimulus. The exact number of flankers was uniformly sampled from the range $[10,25]$.
The length of each line segment varied between $32$ and \SI{64}{px}.
For the flankers consisting of two line segments, both lines had the same length and the angle between the line segments was at least \ang{45}.
We added the flankers successively to the image and thereby ensured a minimal distance of \SI{10}{px} between the line centers.
To ensure that the corresponding image pairs would have the same flankers, the distances to both the closed and open version of the main contour were accounted for when re-sampling flankers.
If a flanker did not fulfill this criterion, a new flanker was sampled of the same size and the same number of line segments, but it was placed somewhere else.
If a flanker extended over the border of the image, the flanker was cropped.

\subsubsection{Line-drawing with Curvy Lines as Main Contour} \label{curvy_desc}
For some of the generalization sets, the contours consisted of curvy instead of straight lines.
These were generated by modulating a circle of a given radius $r_c$ with a radial frequency function that was defined by two sinusoidal functions.
The radius of the contour was thus given by
\begin{equation}
    r(\phi) = A_1 \sin(f_1 (\phi + \theta_1)) + A_2 \sin(f_2 (\phi + \theta_2)) + r_c, 
\end{equation}
with the frequencies $f_1$ and $f_2$, (integers between $1$ and $6$), amplitudes $A_1$ and $A_2$ (random values between $15$ and $45$) and phases $\theta_1$ and $\theta_2$ (between $0$ and $2\pi$).
Unless stated otherwise, the diameter (diameter = $2 \times r_c$) was a random value between $50$ and \SI{100}{px}, and the contour was positioned in the center of the image. 
The open contours were obtained by removing a circular segment of size $\phi_o = \frac{\pi}{3}$ at a random phase (see Figure \ref{fig:cc_methods}B).

For two of the generalization data sets we used dashed contours which were obtained by masking out 20 equally distributed circular segments each of size $\phi_d = \frac{\pi}{20}$.

\subsubsection{Details on Generalization Data Sets}
We constructed $15$ variants of the data set to test generalization performance. Nine variants consisted of contours with straight lines. Six of these featured varying line styles like changes in line width ($10$, $13$, $14$) and/or line color ($11$, $12$). For one variant ($5$), we increased the number of edges in the main contour. Another variant ($4$) had no flankers, and yet another variant ($6$) featured asymmetric flankers. For variant $9$, the lines were binarized (only black or gray pixels instead of different gray tones).

In another six variants, the contours as well as the flankers were curved, meaning that we modulated a circle with a radial frequency function. The first four variants did not contain any flankers and the main contour had a fixed size of \SI{50}{px} ($3$), \SI{100}{px} ($1$) and \SI{150}{px} ($8$). For another variant ($15$), the contour was a dashed line. Finally, we tested the effect of different flankers by adding one additional closed, yet dashed contour ($2$) or one to four open contours ($7$).

Below, we provide more details on some of these data sets:

\textbf{Black-White-Black lines ($12$).} For all contours, black lines enclosed a white one in the middle. Each of these three lines had a thickness of \SI{1.5}{px} which resulted in a total thickness of \SI{4.5}{px}.

\textbf{Asymmetric flankers ($6$).} The two-line flankers consisted of one long and one short line instead of two equally long lines. 

\textbf{W/ dashed flanker ($2$).} This data set with curvy contours contained an additional dashed, yet closed contour as a flanker. 
It was produced like the main contour in the dashed main contour set. 
To avoid overlap of the contours, the main contour and the flanker could only appear at four determined positions in the image, namely the corners.

\textbf{W/ multiple flankers ($7$).} In addition to the curvy main contour, between one and four open curvy contours were added as flankers.
The flankers were generated by the same process as the main contour.
The circles that were modulated had a diameter of \SI{50}{px} and could appear at either one of the four corners of the image or in the center.

\subsection{Psychophysical Experiment} \label{psychopyhysics_cc}
To estimate how well humans would be able to distinguish closed and open stimuli, we performed a psychophysical experiment in which observers reported which of two sequentially presented images contained a closed contour (two-interval forced choice (``2-IFC'') task).

\begin{figure}
 \centering 
 \includegraphics[width=\linewidth]{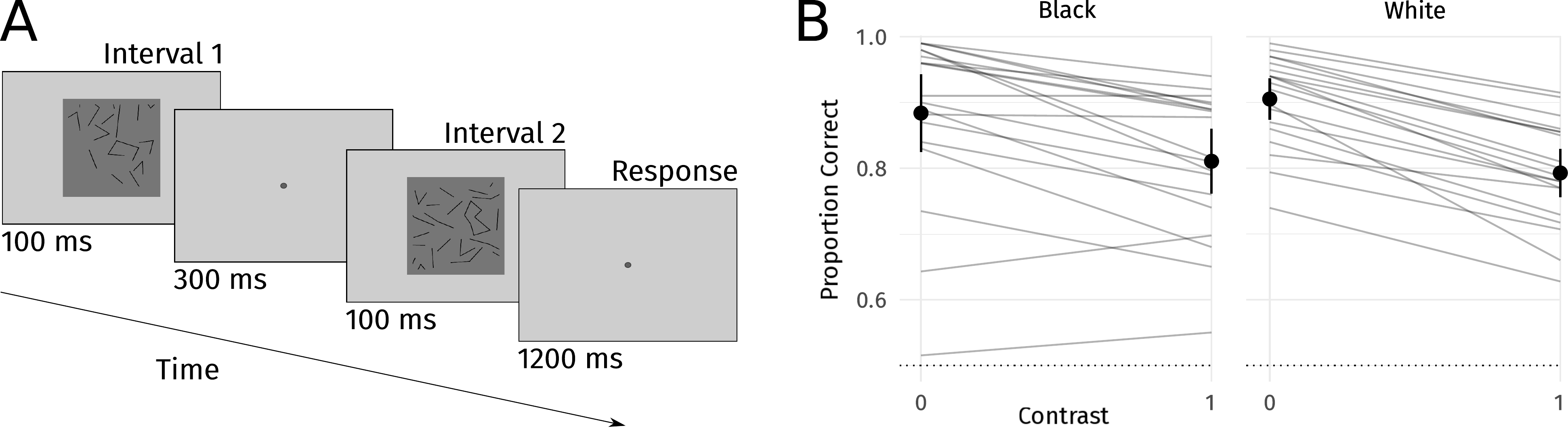}
 \caption{A: In a 2-IFC task, human observers had to tell which of two images contained a closed contour. B: Accuracy of the 20 na\"{i}ve observers for the different conditions.} 
 \label{fig:cc_human_exp}
\end{figure}

\subsubsection{Stimuli}
The images of the closed contour data set were used as stimuli for the psychophysical experiments.
Specifically, we used the images from the test sets that were used to evaluate the performance of the models.
For our psychophysical experiments, we used two different conditions: the images contained either black (i.i.d. to the training set) or white contour lines. The latter was one one of the generalization test sets.

\subsubsection{Apparatus} 
Stimuli were displayed on a VIEWPixx 3D LCD (VPIXX Technologies; spatial resolution $1920 \times 1080$ px, temporal resolution \SI{120}{Hz}, operating with the scanning backlight turned off).
Outside the stimulus image, the monitor was set to mean gray.
Observers viewed the display from \SI{60}{cm} (maintained via a chinrest) in a darkened chamber.
At this distance, pixels subtended approximately \ang{0.024} degrees on average (\SI{41}{ps} per degree of visual angle).
The monitor was linearized (maximum luminance \SI{260}{\mathrm{cd}/ \mathrm{m}^2} using a Konica-Minolta LS-100 photometer.
Stimulus presentation and data collection was controlled via a desktop computer (Intel Core i5-4460 CPU, AMD Radeon R9 380 GPU) running Ubuntu Linux (16.04 LTS), using the Psychtoolbox Library \citep[][version 3.0.12]{pelli1997videotoolbox,kleiner2007s,brainard1997psychophysics} and the iShow library (\url{http://dx.doi.org/10.5281/zenodo.34217}) under MATLAB (The Mathworks, Inc., R2015b).

\subsubsection{Participants}
In total, $19$ na\"{i}ve observers ($4$ male, $15$ female, age: $25.05$ years, SD = $3.52$) participated in the experiment. Observers were paid $10$ \euro per hour for participation. Before the experiment, all subjects had given written informed consent for participating. All subjects had normal or corrected to normal vision. All procedures conformed to Standard 8 of the American Psychological 405 Association’s “Ethical Principles of Psychologists and Code of Conduct” (2010).

\subsubsection{Procedure}
On each trial, one closed and one open contour stimulus were presented to the observer (see Figure \ref{fig:cc_human_exp} A).
The images used for each trial were randomly picked, but we ensured that the open and closed images shown in the same trial were not the ones that were almost identical to each other (see "Generation of Image Pairs" in Appendix \ref{appendix_pairs}).
Thus, the number of edges of the main contour could differ between the two images shown in the same trial.
Each image was shown for \SI{100}{ms}, separated by a \SI{300}{ms} inter-stimulus interval (blank gray screen).
We instructed the observer to look at the fixation spot in the center of the screen.
The observer was asked to identify whether the image containing a closed contour appeared first or second.
The observer had \SI{1200}{ms} to respond and was given feedback after each trial.
The inter-trial interval was \SI{1000}{ms}.
Each block consisted of $100$ trials and observers performed five blocks.
Trials with different line colors and varying background images (contrasts including $0$, $0.4$ and $1$) were blocked.
Here, we only report the results for black and white lines of contrast $0$.
Upon the first time that a block with a new line color was shown, observers performed a practice session with $48$ trials of the corresponding line color.

\subsection{Training of ResNet-50 model}
We fine-tuned a ResNet-50 \citep{he2016deep} pre-trained on ImageNet \citep{imagenet_cvpr09}, on the closed contour task. We replaced the last fully connected, $1000$-way classification layer by a layer with only one output neuron to perform binary classification with a decision threshold of $0$. The weights of all layers were fine-tuned using the optimizer Adam \citep{kingma2014adam} with a batch size of $64$. All images were pre-processed to have the same mean and standard deviation and were randomly mirrored horizontally and vertically for data augmentation. The model was trained on $14,000$ images for $10$ epochs with a learning rate of $0.0003$. We used a validation set of $5,600$ images.

\paragraph{Generalization Tests} To determine the generalization performance, we evaluated the model on the test sets without any further training.
Each of the test sets contained $5,600$ images.
Poor accuracy could simply result from a sub-optimal decision criterion rather than because the network would not be able to tell the stimuli apart.
To account for the distribution shift between the original training images and the generalization tasks, we optimized the decision threshold (a single scalar) for each data set.
To find the optimal threshold for each data set, we subdivided the interval, in which $95\%$ of all logits lie, into $100$ sub points and picked the threshold that would lead to the highest performance.

\subsection{Training of BagNet-33 model}
To test an alternative decision-making mechanism to global contour integration, we trained and tested a BagNet-33 \citep{brendel2019approximating} on the closed contour task. Like the ResNet-50 model, it was pre-trained on ImageNet \citep{imagenet_cvpr09} and we replaced the last fully connected, $1000$-way classification layer by a layer with only one output neuron. We fine-tuned the weights using the optimizer AdaBound \citep{luo2019adaptive} with an initial and final learning rate of $0.0001$ and $0.1$, respectively. The training images were generated on-the-fly, which meant that new images were produced for each epoch. In total, the fine-tuning lasted $100$ epochs and we picked the weights from the epoch with highest performance.

\begin{figure}
 \centering 
 \includegraphics[width=\linewidth]{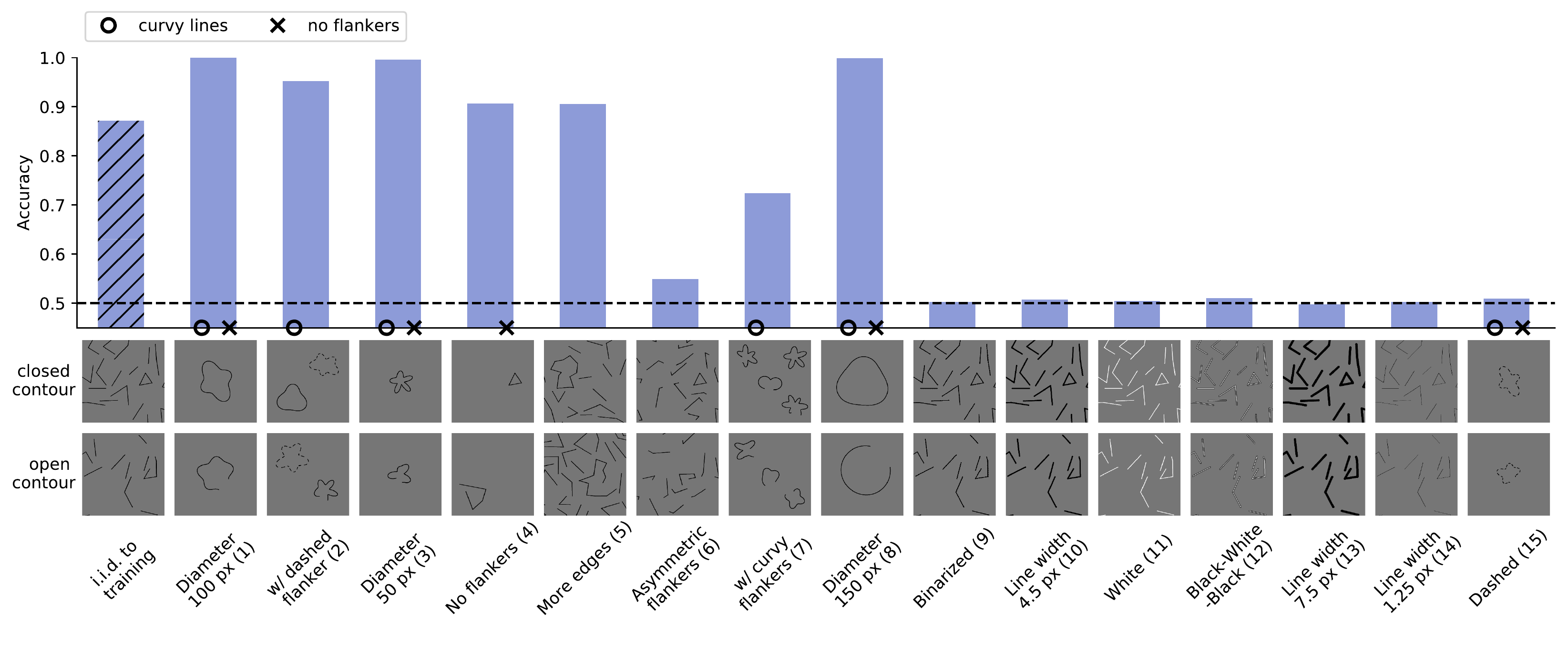}
 \caption{Generalization performances of BagNet-33.}
 \label{fig:cc_bagnet_gen}
\end{figure}

\paragraph{Generalization Tests} The generalization tests were conducted equivalently to the ones with ResNet-50. The results are shown in Figure \ref{fig:cc_bagnet_gen}.

\subsection{Additional Experiment: Increasing the Task Difficulty by Adding a Background Image} \label{random_contrast}
We performed an additional experiment, where we tested if the model would become more robust and thus generalized better if we trained on a more difficult task.
This was achieved by adding an image to the background, such that the model had to learn how to separate the lines from the task-irrelevant background.

\begin{figure}
 \centering 
 \includegraphics[width=\linewidth]{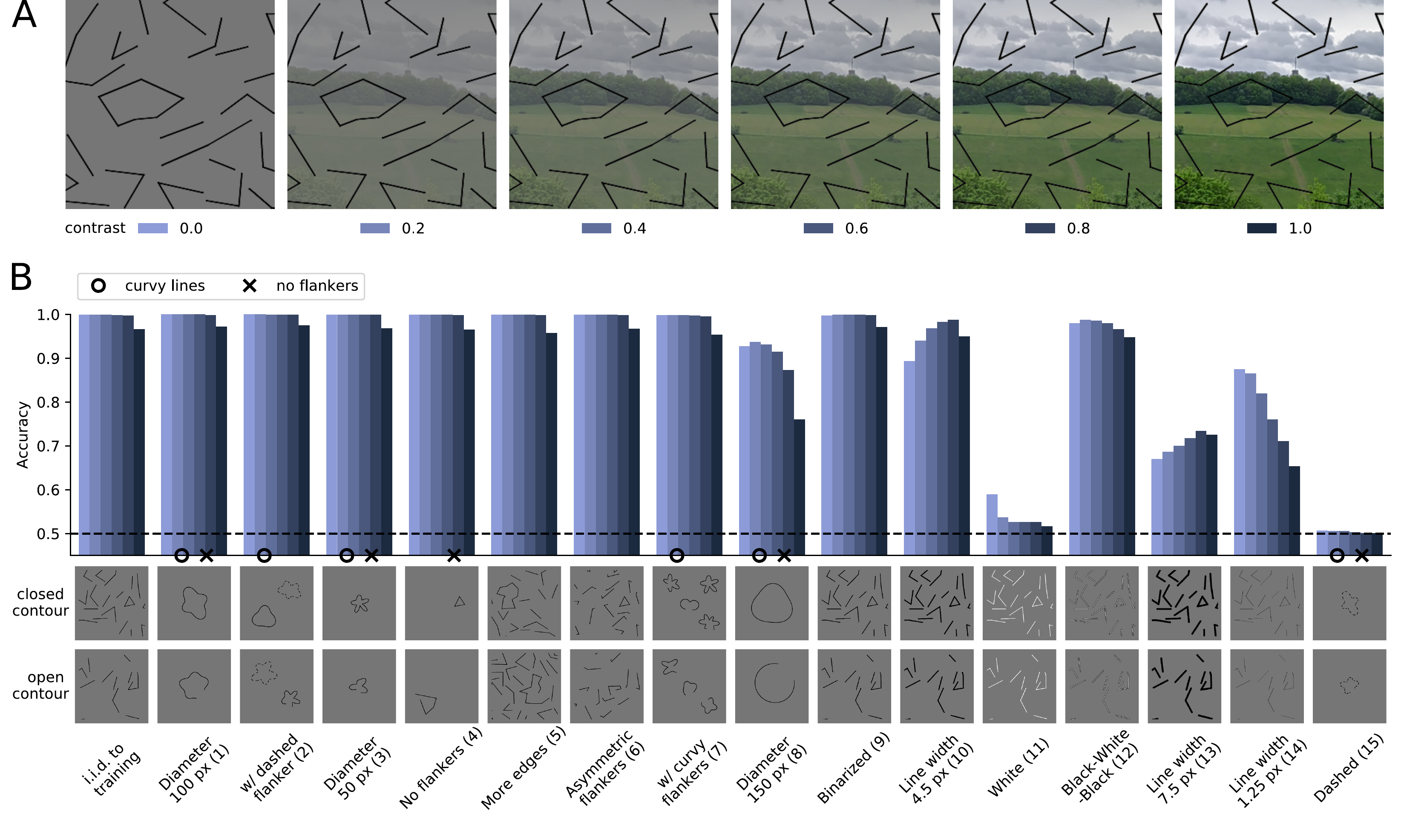}
 \caption{A: An image of varying contrast was added as background. B: Generalization performances of our models trained on random contrast levels and tested on single contrast levels.}
 \label{fig:cc_randomcontrast}
\end{figure}

In our experiment, we fine-tuned our ResNet-50-based model on images with a background image of a uniformly sampled contrast.
For each data set, we evaluated the model separately on six discrete contrast levels \{0, 0.2, 0.4, 0.6, 0.8, 1\} (see Figure \ref{fig:cc_randomcontrast}A).
We found that the generalization performance varied for some data sets compared to the experiment in the main body (see Figure \ref{fig:cc_randomcontrast}B). 

\section{SVRT}
\label{appendix_svrt}

\subsection{Methods}
\paragraph{Data set}
We used the original C-code provided by \citet{fleuret2011comparing} to generate the images of the SVRT data set.
The images had a size of $128 \times 128$ pixels.
For each problem, we used up to $28,000$ images for training, $5,600$ images for validation and $11,200$ images for testing.

\paragraph{Experimental Procedures} \label{exp_procedure_SVRT}
For each of the SVRT problems, we fine-tuned a ResNet-50 that was pretrained on ImageNet \citep{imagenet_cvpr09} (as described in section \ref{resnet_cc}). 
The same pre-processing, data augmentation, optimizer and batch size as for the closed contour task were used.

For the different experiments, we varied the number of training images. We used subsets containing either $28,000$, $1000$ or $100$ images.
The number of epochs depended on the size of the training set: The model was fine-tuned for respectively $10$, $280$ or $2800$ epochs.
For each training set size and SVRT problem, we used the best learning rate after a hyper-parameter search on the validation set, where we tested the learning rates [\num{6e-5}, \num{1e-4}, \num{3e-4}].

As a control experiment, we also initialized the model with random weights and we again performed a hyper-parameter search over the learning rates [\num{3e-4}, \num{6e-4}, \num{1e-3}].

\subsection{Results} \label{results_svrt}
\begin{figure}
 \centering 
 \includegraphics[width=\linewidth]{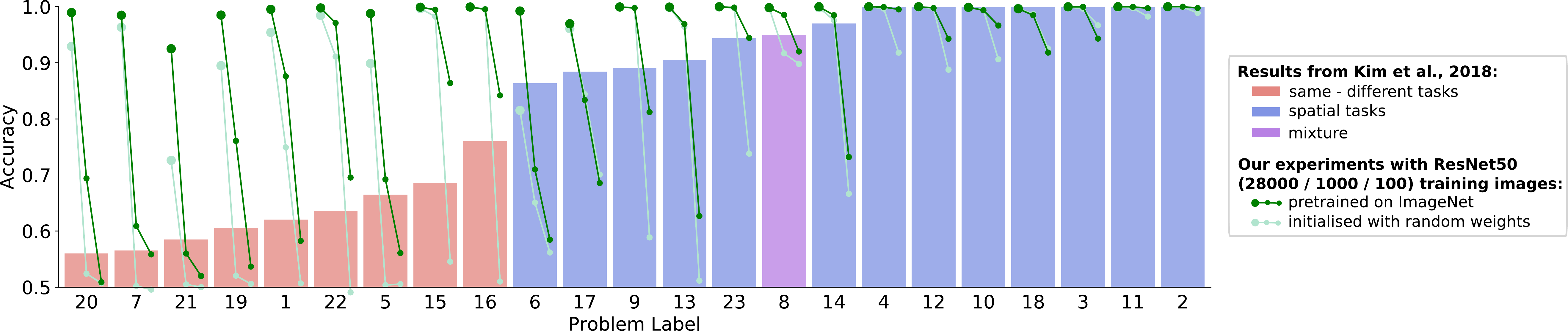}
 \caption{Accuracy of the models for the individual problems. Problem 8 is a mixture of same-different task and spatial task. In Figure \ref{fig:svrt_main} this problem was assigned to the spatial tasks. Bars re-plotted from \citet{kim2018not}.}
 \label{fig:svrt_appendix}
\end{figure}

In Figure \ref{fig:svrt_appendix} we show the results for the individual problems. 
When using $28,000$ training images, we reached above $90$\% accuracy for all SVRT problems, including the ones that required same-different judgments (see also Figure \ref{fig:svrt_main}B).
When using less training images the performance on the test set was reduced. In particular, we found that the performance on same-different tasks dropped more rapidly than on spatial reasoning tasks. 
If the ResNet-50 was trained from scratch (i.e. weights were randomly initialized instead of loaded from pre-training on ImageNet), the performance dropped only slightly on all but one spatial reasoning task. Larger drops were found on same-different tasks.

\section{Recognition Gap}

\subsection{Details on Methods}\label{rec_gap:methods}

\paragraph{Data set}
We used two data sets for this experiment. One consisted of ten natural, color images whose grayscale versions were also used in the original study by \citet{ullman2016atoms}. We discarded one image from the original data set as it does not correspond to any ImageNet class. For our ground truth class selection, please see Table \ref{fig:ImageNet_classes}. The second data set consisted of $1000$ images from the ImageNet \citep{imagenet_cvpr09} validation set. All images were pre-processed like in standard training of ResNet (i.e. resizing to 256 $ \times 256$ pixels, cropping centrally to $224 \times 224$ pixels and normalizing). 

\paragraph{Model}
In order to evaluate the recognition gap, the model had to be able to handle small input images. Standard networks like ResNet \citep{he2016deep} are not equipped to handle small images. In contrast, BagNet-33 \citep{brendel2019approximating} allows to straightforwardly analyze images as small as $33 \times 33$ pixels and hence was our model of choice for this experiment. It is a variation of ResNet-50 \citep{he2016deep} where most $3 \times 3$ kernels are replaced by $1 \times 1$ kernels such that the receptive field size at the top-most convolutional layer is restricted to $33 \times 33$ pixels.

\paragraph{Machine-Based Search Procedure for Minimal Recognizable Images} \label{exp+model_recgap} Similar to \citet{ullman2016atoms}, we defined minimal recognizable images or configurations (MIRCs) as those patches of an image for which an observer - by which we mean an ensemble of humans or one or several machine algorithms - reaches $\geq 50\%$ accuracy, but any additional $20\%$ cropping of the corners or $20\%$ reduction in resolution would lead to an accuracy $< 50\%$. MIRCs are thus inherently observer-dependent. The original study only searched for MIRCs in humans. We implemented the following procedure to find MIRCs in our DNN: We passed each pre-processed image through BagNet-33 and selected the most predictive crop according to its probability. See Appendix \ref{rec_gap:custom_probability} on how to handle cases where the probability saturates at $100\%$ and Appendix \ref{rec_gap:class_stride_analysis} for different treatments of ground truth class selections. If this probability of the full-size image for the ground-truth class was $\geq 50\%$, we again searched for the $80\%$ subpatch with the highest probability. We repeated the search procedure until the class probability for all subpatches fell below $50\%$. If the $80\%$ subpatches would be smaller than $33 \times 33$ pixels, which is BagNet-33's smallest natural patch size, the crop was increased to $33 \times 33$ pixels using bilinear sampling. We evaluated the recognition gap as the difference in accuracy between the MIRC and the \textit{best-performing} sub-MIRC. This definition was more conservative than the one from \citet{ullman2016atoms} who considered the maximum difference between a MIRC and its sub-MIRCs, i.e. the difference between the MIRC and the \textit{worst-performing} sub-MIRC. Please note that one difference between our machine procedure and the psychophysics experiment by \citet{ullman2016atoms} remained: The former was greedy, whereas the latter corresponded to an exhaustive search under certain assumptions.

\subsection{Analysis of Different Class Selections and Different Number of Descendants}
\label{rec_gap:class_stride_analysis}

Treating the ten stimuli from \citet{ullman2016atoms} in our machine algorithm setting required two design choices: We needed to both pick suitable ground truth classes from ImageNet for each stimulus as well as choose if and how to combine them. The former is subjective and using relationships from WordNet Hierarchy \citep{miller1995wordnet} (as \citet{ullman2016atoms} did in their psychophysics experiment) only provides limited guidance. We picked classes to our best judgement (for our final ground truth class choices, please see Table \ref{fig:ImageNet_classes}). Regarding the aspect of handling several ground truth classes, we extended our experiments: We tested whether considering all classes as one ('joint classes', i.e. summing the probabilities) or separately ('separate classes', i.e. rerunning the stimuli for each ground truth class) would have an effect on the recognition gap. As another check, we investigated whether the number of descendant options would alter the recognition gap: Instead of only considering the four corner crops as in the psychophysics experiment by \citet{ullman2016atoms} ('Ullman4'), we looked at every crop shifted by one pixel as a potential new parent ('stride-1'). The results reported in the main body correspond to joint classes and corner crops. Finally, besides analyzing the recognition gap, we also analyzed the sizes of MIRCs and the fractions of images that possess MIRCs for the mentioned conditions.

\begin{figure}
 \centering 
 \includegraphics[width=\linewidth]{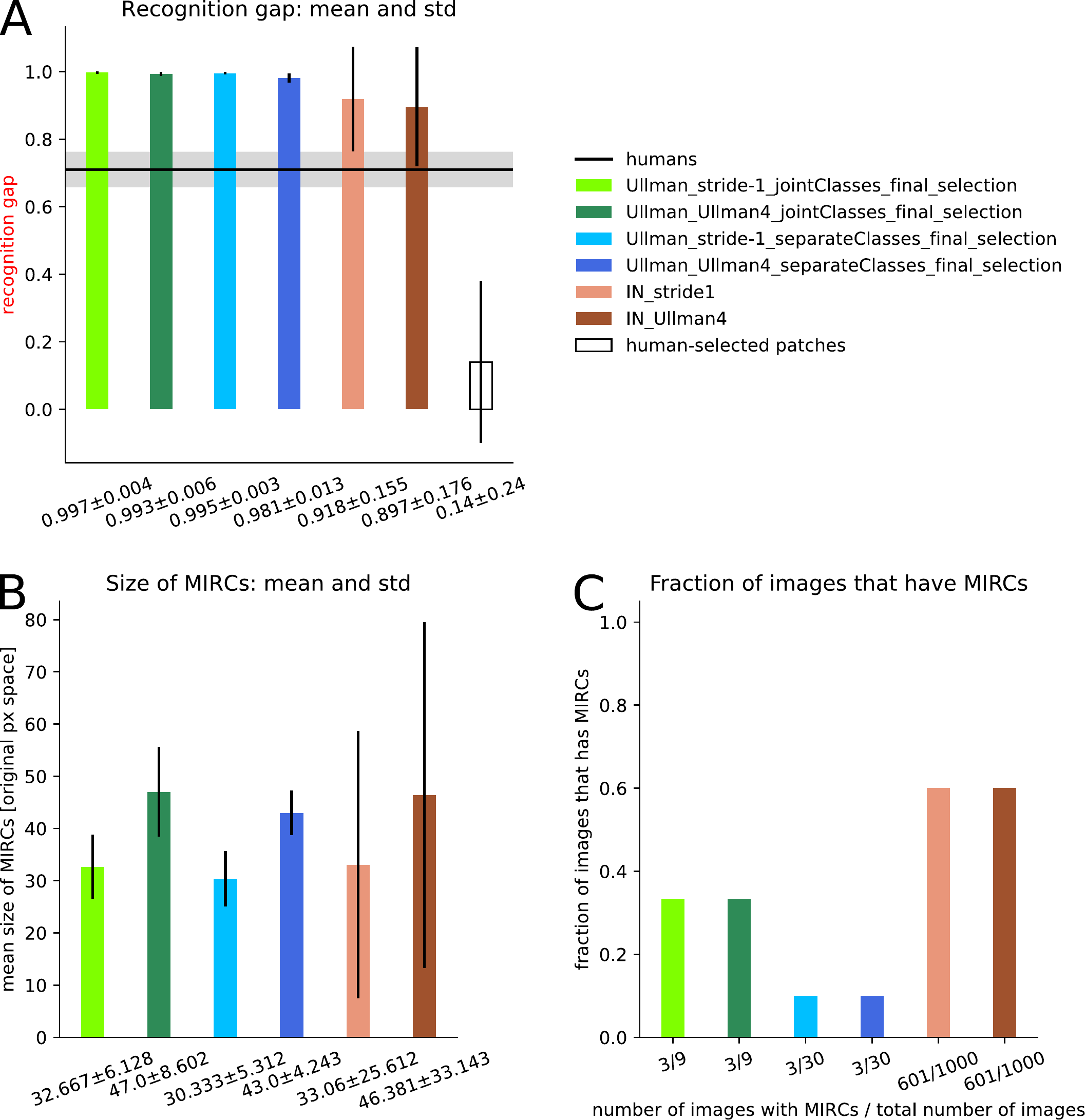}
 \caption{A: Recognition gaps. The legend holds for all subplots. B: Size of MIRCs. C: Fraction of images with MIRCs.}
 \label{fig:recgap_sup}
\end{figure}

Figure \ref{fig:recgap_sup}A shows that all options result in similar values for the recognition gap. The trend of smaller MIRC sizes for stride-1 compared to four corner crops shows that the search algorithm can find even smaller MIRCs when all crops are possible descendants (see. Figure \ref{fig:recgap_sup}B). The final analysis of how many images possess MIRCs (see Figure \ref{fig:recgap_sup}C) shows that recognition gaps only exist for fractions of the tested images: In the case of the stimuli from \citet{ullman2016atoms} three out of nine images, and in the case of ImageNet about $60\%$ of the images have MIRCs. This means that the recognition performance of the initial full-size configurations was $\geq 50\%$ for those fractions only. Please note that we did not evaluate the recognition gap over images that did not meet this criterion. In contrast, \citet{ullman2016atoms} average only across MIRCs that have a recognition rate above $65\%$ and sub-MIRCs that have a recognition rate below $20\%$ (personal communication, 2019). The reason why our model could only reliably classify three out of the nine stimuli from \citep{ullman2016atoms} can partly be traced back to the oversimplification of single-class-attribution in ImageNet as well as to the overconfidence of deep learning classification algorithms \citep{guo2017calibration}: They often attribute a lot of evidence to one class, and the remaining ones only share very little evidence.

\subsection{Selecting Best Crop when Probabilities Saturate}\label{rec_gap:custom_probability}

We observed that several crops had very high probabilities and therefore used the ``logit''-measure $logit(p)$, where $p$ is the probability. It is defined as the following: $logit(p) = log(\frac{p}{1-p})$. Note that this measure is different from what the deep learning community usually refers to as ``logits'', which are the values before the softmax-layer. In the following, we denote the latter values as $\mathbf{z}$. The logit $logit(p)$ is monotonic w.r.t. to the probability $p$, meaning that the higher the probability $p$, the higher the logit $logit(p)$. However, while $p$ saturates at $100\%$, $logit(p)$ is unbounded. Therefore, it yields a more sensitive discrimination measure between image patches $j$ that all have $p(\mathbf{z}^j) = 1$, where the superscript $j$ denotes different patches.

In the following, we will provide a short derivation for the logit $logit(p)$. Consider a single patch with the correct class $c$. We start with the probability $p_c$ of class $c$, which can be obtained by plugging the logits $z_i$ into the softmax-formula, where $i$ corresponds to the classes $[0, ..., 1000]$.

\begin{equation}
p_c(\mathbf{z}) = \frac{exp(z_c)}{exp(z_c) + \sum\limits_{i \neq c} exp(z_i)} 
\end{equation}

Since we are interested in the probability of the correct class, it holds that $p_c(\mathbf{z}) \neq 0 $. Thus, in the regime of interest, we can invert both sides of the equation. After simplifying, we get:

\begin{equation}
\frac {1}{p_c(\mathbf{z})} -1= \frac{ \sum\limits_{i \neq c} exp(z_i)} {exp(z_c)}.
\end{equation}

When taking the negative logarithm on both sides, we obtain:
\begin{align}
&\Leftrightarrow &-log \left(\frac {1}{p_c(\mathbf{z})} -1\right) & = -log \left(\frac{ \sum\limits_{i \neq c} exp(z_i)} {exp(z_c)}\right)\\
&\Leftrightarrow &-log \left( \frac{1-p_c(\mathbf{z})}{p_c(\mathbf{z})} \right) & = -log \left(\sum\limits_{i \neq c} exp(z_i)\right)  - \left(-log(exp(z_c))\right)\\
&\Leftrightarrow &log \left( \frac{p_c(\mathbf{z})}{1-p_c(\mathbf{z})}  \right) & = z_c -log \left(\sum\limits_{i \neq c} exp(z_i)\right) 
\end{align}

The left-hand side of the equation is exactly the definition of the logit $logit(p)$. Intuitively, it measures in log-space how much the network's belief in the correct class outweighs the belief in all other classes taken together. The following reassembling operations illustrate this:

\begin{equation}
\begin{split}
logit(p_c) & = log \left( \frac{p_c(\mathbf{z})}{1-p_c(\mathbf{z})}  \right)\\
& = \underbrace{log \bigl( p_c(\mathbf{z}) \bigr)}_\text{log probability of correct class} - 
\underbrace{log \bigl( 1-p_c(\mathbf{z}) \bigr)}_\text{log probability of all incorrect classes}
\end{split}
\end{equation}

The above formulations regarding one correct class hold when adjusting the experimental design to accept several classes $k$ as correct predictions. In brief, the logit $logit(p_C(z))$, where $C$ stands for several classes, then states:

\begin{equation}
\begin{split}
    logit(p_C(\mathbf{z}))
    & = -log \left(  \frac{1}{p_{c_1}(\mathbf{z}) + p_{c_2}(\mathbf{z}) + ... + p_{c_k}(\mathbf{z})} - 1 \right) \\
    & = -log \left(  \frac{1}{\sum\limits_{k}p_k(\mathbf{z})} - 1 \right) \\
    & = \underbrace{log \bigl( \sum\limits_{k}p_k(\mathbf{z}) \bigr)}_\text{log probability of all correct classes}  - 
        \underbrace{log \bigl( 1-\sum\limits_{k}p_k(\mathbf{z}) \bigr)}_\text{log probability of all incorrect classes}\\
    & = log \bigl( \sum\limits_{k}exp(z_k)  \bigr) -
        log \bigl( \sum\limits_{i \neq k}exp(z_i) \bigr)
\end{split}
\end{equation}

\subsection{Selection of ImageNet Classes for Stimuli of Ullman et al. (2016)}\label{ImageNet_classes}

Note that our selection of classes is different from the one used by \citet{ullman2016atoms}. We went through all classes for each image and selected the ones that we considered sensible. The tenth image of the eye does not have a sensible ImageNet class, hence only nine stimuli from \citet{ullman2016atoms} are listed in Table \ref{fig:ImageNet_classes}.

\begin{table}[]
\begin{tabular}{|l|l|l|l|}
\hline
Image   & \begin{tabular}[c]{@{}l@{}} WordNet \\ Hierarchy ID \end{tabular} & WordNet Hierarchy description                                                                                                  & \begin{tabular}[c]{@{}l@{}}Neuron number in ResNet-50\\ (indexing starts at 0)\end{tabular} \\ \hline
fly     & n02190166            & fly                                                                                                                            & 308                                                                                         \\ \hline
ship    & n02687172            & \begin{tabular}[c]{@{}l@{}}aircraft carrier, carrier, flattop, attack \\ aircraft carrier \end{tabular}                                                                     & 403                                                                                         \\ \hline
        & n03095699            & \begin{tabular}[c]{@{}l@{}} container ship, containership, container \\ vessel \end{tabular}                                                                                  & 510                                                                                         \\ \hline
        & n03344393            & fireboat                                                                                                                       & 554                                                                                         \\ \hline
        & n03662601            & lifeboat                                                                                                                       & 625                                                                                         \\ \hline
        & n03673027            & liner, ocean liner                                                                                                             & 628                                                                                         \\ \hline
eagle   & n01608432            & kite                                                                                                                           & 21                                                                                          \\ \hline
        & n01614925            & \begin{tabular}[c]{@{}l@{}}bald eagle, American eagle, Haliaeetus \\ leucocephalus\end{tabular}                                & 22                                                                                          \\ \hline
glasses & n04355933            & sunglass                                                                                                                       & 836                                                                                         \\ \hline
        & n04356056            & sunglasses, dark glasses, shades                                                                                               & 837                                                                                         \\ \hline
bike    & n02835271            & \begin{tabular}[c]{@{}l@{}} bicycle-built-for-two, tandem bicycle, \\ tandem\end{tabular}                                                                                     & 444                                                                                         \\ \hline
        & n03599486            & jinrikisha, ricksha, rickshaw                                                                                                  & 612                                                                                         \\ \hline
        & n03785016            & moped                                                                                                                          & 665                                                                                         \\ \hline
        & n03792782            & mountain bike, all-terrain bike, off-roader                                                                                    & 671                                                                                         \\ \hline
        & n04482393            & tricycle, trike, velocipede                                                                                                    & 870                                                                                         \\ \hline
suit    & n04350905            & suit, suit of clothes                                                                                                          & 834                                                                                         \\ \hline
        & n04591157            & windsor tie                                                                                                                    & 906                                                                                         \\ \hline
plane   & n02690373            & airliner                                                                                                                       & 404                                                                                         \\ \hline
horse   & n02389026            & sorrel                                                                                                                         & 339                                                                                         \\ \hline
        & n03538406            & horse cart, horse-cart                                                                                                         & 603                                                                                         \\ \hline
car     & n02701002            & ambulance                                                                                                                      & 407                                                                                         \\ \hline
        & n02814533            & \begin{tabular}[c]{@{}l@{}}beach wagon, station wagon, wagon \\ estate car, beach waggon, station waggon, \\ waggon\end{tabular} & 436                                                                                         \\ \hline
        & n02930766            & cab, hack, taxi, taxicab                                                                                                       & 468                                                                                         \\ \hline
        & n03100240            & convertible                                                                                                                    & 511                                                                                         \\ \hline
        & n03594945            & jeep, landrover                                                                                                                & 609                                                                                         \\ \hline
        & n03670208            & limousine, limo                                                                                                                & 627                                                                                         \\ \hline
        & n03769881            & minibus                                                                                                                        & 654                                                                                         \\ \hline
        & n03770679            & minivan                                                                                                                        & 656                                                                                         \\ \hline
        & n04037443            & racer, race car, racing car                                                                                                    & 751                                                                                         \\ \hline
        & n04285008            & sports car, sport car                                                                                                          & 817                                                                                         \\ \hline
\end{tabular}
\caption {Selection of ImageNet Classes for Stimuli of \citet{ullman2016atoms}}
\label{fig:ImageNet_classes}
\end{table}

\end{appendices}

\end{document}